\documentclass[sigconf]{acmart}
\usepackage{hyperref}
\usepackage{algorithm}
\usepackage[noend]{algpseudocode}
\usepackage{graphicx}
\usepackage{textcomp}
\usepackage{xcolor}

\setcopyright{none}
\renewcommand\footnotetextcopyrightpermission[1]{}

\setcopyright{acmcopyright}
\copyrightyear{2018}
\acmYear{2018}
\acmDOI{10.1145/1122445.1122456}

\acmConference[Woodstock '18]{Woodstock '18: ACM Symposium on Neural
  Gaze Detection}{June 03--05, 2018}{Woodstock, NY}
\acmBooktitle{Woodstock '18: ACM Symposium on Neural Gaze Detection,
  June 03--05, 2018, Woodstock, NY}
\acmPrice{15.00}
\acmISBN{978-1-4503-XXXX-X/18/06}

\begin{document}

\title{Accelerating Gradient-based Meta Learner}

\author{Varad Pimpalkhute}
\affiliation{%
  \institution{IIIT Nagpur}
  \country{India}
}
\email{pimpalkhutevarad@gmail.com}

\author{Amey Pandit, Mayank Mishra, Rekha Singhal}
\affiliation{%
  \institution{TCS Research} 
  \country{India}
}
\email{amey.pandit@tcs.com, mishra.m@tcs.com, rekha.singhal@tcs.com}





\begin{abstract}
	Meta Learning has been in focus in recent years due to the meta-learner model's ability to adapt well and generalize to new tasks, thus, reducing both the time and data requirements for learning. However, a major drawback of meta learner is that, to reach to a state from where learning new tasks becomes feasible with less data, it requires a large number of iterations and a lot of time. We address this issue by proposing various acceleration techniques to speed up meta learning algorithms such as MAML (Model Agnostic Meta Learning). We present 3.73X acceleration on a well known RNN optimizer based meta learner proposed in literature ~\cite{DBLP:conf/icml/WichrowskaMHCDF17}. We introduce a novel method of training tasks in clusters, which not only accelerates the meta learning process but also improves model accuracy performance.
\end{abstract}





\maketitle
\pagestyle{plain}

\section{Introduction}
Artificial general intelligence (AGI) is a hypothetical concept where machines are capable of learning and thinking like humans. Meta learning also known as learning to learn, brings us one step closer to achieve AGI by learning on a task in a manner similar to humans. To give an analogy, in Meta-learning paradigm, a distribution of related tasks (say racket sports: Badminton, Tennis etc.) are used to train a model (human) which can use this "experience" to learn any new task (Squash) with lesser amount of data (lesser practice of Squash than a person who doesn't know any racket sport) and in lesser time.

The meta learning algorithms, like MAML \cite{DBLP:journals/corr/FinnAL17} (and others  ~\cite{DBLP:conf/iclr/RaghuRBV20, DBLP:conf/icml/WichrowskaMHCDF17, DBLP:journals/corr/SnellSZ17}), are time and compute intensive. These algorithms require a lot of iterations over the available tasks (60000 ~\cite{DBLP:journals/corr/FinnAL17}) to make a model converge and be ready to learn a new task quickly with good accuracy.



In this paper, we present our ongoing work involving several performance optimization techniques to reduce the training time of metalearning algorithms. We will discuss these techniques in the context of acceleration of a gradient  based meta learner proposed in ~\cite{DBLP:conf/icml/WichrowskaMHCDF17}. A gradient based meta learner learns how gradients change during gradient descent based optimizations of various learning tasks (learning task involves training a model). The goal is to enable rapid model parameter updates (faster than optimizers like SGD, ADAM, RMSProp) after each learning iteration \footnote{There are other approaches of meta learning too, namely, metric based and loss based meta learners \cite{hospedales2020meta}.}. 

\begin{figure}
	\centering
	\includegraphics[width = 3.2in]{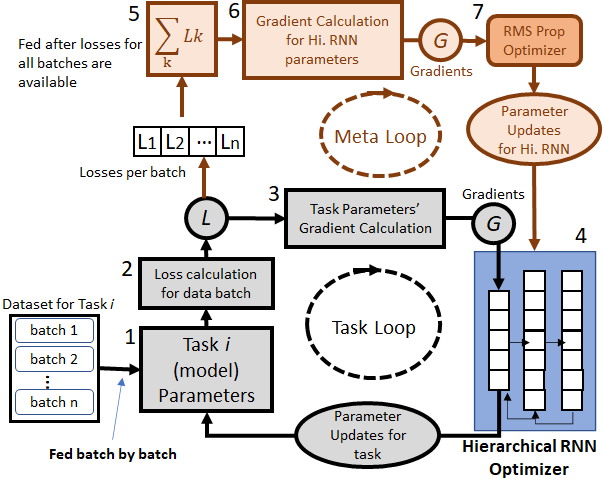}
	\caption{Meta-training on a single task using a Hierarchical RNN-based optimizer.}
	\label{fig:rnn_optimizer}
\end{figure}

The gradient based meta learner which is our focus in this paper employs a Hierarchical RNN \footnote{RNN: Recurrent Neural Network} architecture to ensembles the loss landscapes of various small tasks. Figure \ref{fig:rnn_optimizer} shows how this meta learner learnt from a single task's training process (the same is repeated for every task). There are two learning loops: \textbf{a) Task loop} - which updates the task parameters for every batch of data using hierarchical RNN as optimizer; and \textbf{b) Meta loop} - which updates the parameters of RNN optimizer itself using loss accumulated across all batches of task learning using a RMSProp optimizer. This process is repeated multiple times (N epochs) for each task and sequentially for every task available in a single meta iteration. Large number of such meta iterations are required to let Hierarchical RNN based meta optimizer converge.

\begin{figure}
	\centering
	\includegraphics[width = 1.9in]{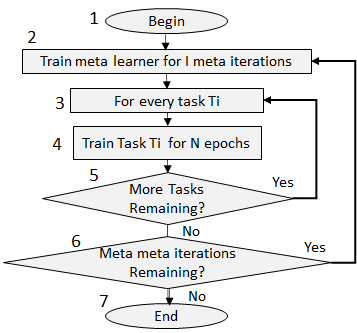}
	\caption{Earlier Proposed approach in the paper}
	\label{fig:vanilla}
\end{figure}

Figure \ref{fig:vanilla} shows the sequential nature of the meta training approach and the iterations involved using a flow-chart. We improved on the model performance by introducing a better and wider meta-optimization pipeline, which incorporates the optimizer parameters being learned in form of task clusters. We achieved a speedup of more than 2x with our proposed approach and overall speedup of 3.73x by also incorporating several coding related optimizations.

The rest of the paper is structured as follows: We briefly discuss the previous related work in the field of meta learning and accelerations in Section \ref{background}. We revisit definition of tasks in Section \ref{tasks_preparation}. We propose the approaches for accelerating Hierarchical RNN in Section \ref{methodology}. We verify and discuss the effectiveness of our experiments in Section \ref{results}. Finally, we discuss the ongoing work in Section \ref{sec:ongoing_future} and conclude the paper in Section \ref{sec:conclusions}.

\section{Related Work} \label{background}

Gradient based meta-learners proposed in literature ~\cite{DBLP:conf/iclr/RaghuRBV20, DBLP:journals/corr/MishraRCA17, Lee_2019_CVPR, Franceschi2018BilevelPF} have worked on improving model accuracy. However, very little work has been done on improving the training time of the meta learner, where meta-learner especially MAML is a time hungry algorithm.

Gradient descent on the meta-parameters ($w$) involves computing second-order derivatives or a simplified and cheaper first order implementation ~\cite{DBLP:journals/corr/FinnAL17, Ravi2017OptimizationAA,  DBLP:journals/corr/abs-1803-02999}. ~\cite{Ravi2017OptimizationAA} proposes ANIL which makes use of feature reuse for few-shot learning. ANIL simplifies on MAML by using almost no inner loop; thus being more computationally efficient than MAML, getting a speed up as high as 1.8x. ~\cite{DBLP:journals/corr/abs-1803-02999} focuses on decreasing the computations and memory by not unrolling a computational graph or calculating any second derivatives. ~\cite{10.1007/978-3-030-61527-7_20} propose a distributed framework to accelerate the learning process of MAML.

In this work, we also aim to accelerate the learning process by proposing various optimization techniques. Contrary to the above approaches (which involve changing the mathematical model or framework), our approach is more intuitive and less complex.


\section{Tasks Preparation for meta learning} \label{tasks_preparation}
The Hierarchical RNN based meta learner work proposed in \cite{DBLP:conf/icml/WichrowskaMHCDF17} which we intend to accelerate is designed to be an optimizer with rapid updates. Consequently, it becomes important that the optimizer should work for any model, may it be CNN, Fully connected, RNN, logistic regression, etc. Thus, optimizer is trained to give rapid updates by training it on various instances of these models. The instance of a model is considered as task. The diversity of tasks helps in generalizing the model well to new tasks. Datasets to train the task (model) are generated by following the approach mentioned in paper.

\section{Proposed Approaches for Acceleration} \label{methodology}
\begin{figure}
	\centering
	\includegraphics[width = 2.8in]{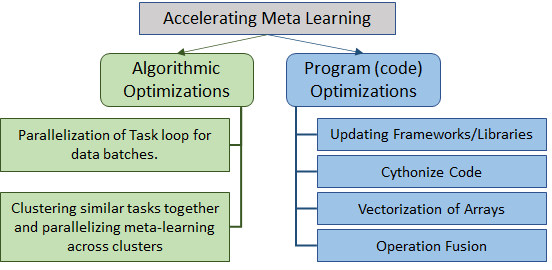}
	\caption{Performance optimizations experimented on RNN optimizer based meta learner}
	\label{fig:optim_tech}
\end{figure}

The approaches we tried for speeding up Hierarchical RNN based meta learner can be divided into two categories A) Algorithmic optimizations, and B) Program optimizations, as shown in Figure \ref{fig:optim_tech}. 

We focus on discussions of Algorithmic optimizations performed in this section, while limiting the discussion around program optimizations tried in experiments section as they are mostly straightforward and simple to understand.

\subsection{Parallelization inside Task Loop}
The task-loop shown in Figures \ref{fig:rnn_optimizer} and \ref{fig:vanilla} involved training over multiple batches of data sequentially. Instead of training over data batches sequentially we train over multiple batches in parallel. We calculate each batch's loss and parameter gradients and then calculate average of individual parameter's gradients. Finally, updating the task parameters using these averaged parameter gradients. The updated model is used in the next iteration. The degree of parallelism is configurable and depends on the underlying hardware.

\subsection{Considering tasks in clusters and training clusters parallely}
Although, meta learning approaches employ similar tasks for training the meta learner to begin with, some tasks in the task set are more similar to each other than other tasks. For example, a softmax regression task will be more similar to other softmax regression tasks in terms of how gradients get updated in each optimisation step than task which involves training a fully connected neural network. The Hierarchical RNN based meta optimizer approach doesn't differentiate between such tasks and while meta training the "sequence" of tasks is randomly chosen.

\begin{figure}[ht]
	\centering
	\includegraphics[width=2.8in]{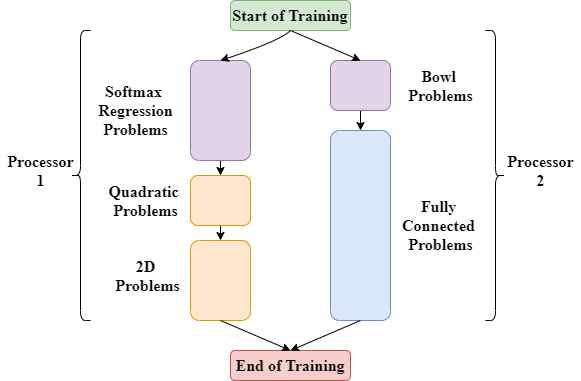}
	\caption{Optimal grouping of task clusters as both the branches are of equal length.}
	\label{fig:dag}
\end{figure}

We grouped the tasks into clusters\footnote{There are various ways to create clusters of tasks. Thus for simplicity, we  consider creation of clusters as a black box which returns k clusters, each cluster having similar tasks.} depending on their similarity with each other and then we performed meta training across these clusters parallely. The Hierarchical RNN optimizer's parameters were updated by using the averaged parameter gradients across these clusters. This iteration of meta training parallely over the clusters is repeated for a number of times (number of meta-iterations is user defined).  To properly utilise the underlying hardware and ensuring that no one branch of parallelization becomes bottleneck and takes too long we group clusters into cluster groups and schedule cluster groups as shown in Figure \ref{fig:dag}. 

The approach is that all cluster-groups should take similar time for meta learning iteration. Figure \ref{fig:dag} shows 5 task clusters containing similar tasks namely Softmax Regression, Quadratic problems, 2D problems, Bowl problems, and Fully connected problems are scheduled as two cluster groups. 

\begin{figure}
	\centering
	\includegraphics[width = 3.3in]{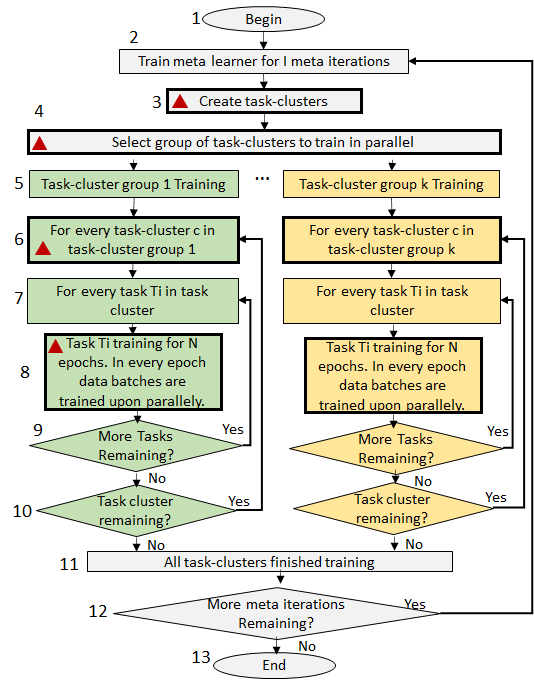}
	\caption{Our Proposed overall approach}
	\label{fig:our}
\end{figure}

Figure \ref{fig:our} gives the complete picture of the algorithmic optimization we performed in form of a flow chart. The original sequential pipeline of meta training mentioned in flowchart of Figure \ref{fig:vanilla} has been broadened by introducing parallelizations in task loop (box 8 in flowchart of Figure \ref{fig:our}) as well as across training task-cluster (green and yellow region). The red triangles indicate the crucial steps where we have placed our optimizations. To summarize our acceleration approach we train all clusters parallely while learning on the tasks in each cluster sequentially. Rest of the procedure is similar to that followed in ~\cite{DBLP:conf/icml/WichrowskaMHCDF17}.

\section{Results \& Discussion} \label{results}
In this section, we perform experiments to verify the effectiveness of proposed acceleration techniques. The hardware and software configuration employed for experiments is mentioned in Table \ref{table:experiment_setup}. Before performing experiments we upgraded the baseline code available from prior work \cite{DBLP:conf/icml/WichrowskaMHCDF17} to newer libraries. 

We performed experiments in a resource constrained setup as can be seen in Table \ref{table:experiment_setup}. We believe that a higher configuration setup will exhibit better acceleration and performance gains due to possibility of higher parallelism.  
All the experiments are performed while ensuring that the achieved test accuracy of 0.65 for test "task-set" (consisting of softmax regression tasks) on base model is matched using our proposed approach. We now present details of experiments performed for each of proposed acceleration technique. 

\begin{center}
	\begin{table}[ht]
		\small
		\caption{Experiment Setup}
		\begin{tabular}{ |p{2.2cm}|p{5cm}|}
			\hline
			\textbf{Type of tasks} & Quadractic, Bowl, Fully Connected, Softmax Regression, 2D Problems\\ \hline
			\textbf{Hardware Setup} & Intel core i5 CPU @2.50GHz, 8GB RAM, 2 physical and 4-logical cores. \\ \hline
			\textbf{Software} & Windows 10 (64-bit) OS, Cython, Tensorflow, Python, Numpy, Pandas, Scikit \\ \hline\end{tabular}
		
		\label{table:experiment_setup}
	\end{table}
\end{center}

\subsection{Acceleration achieved due to parallelism introduced}
Table \ref{table:cluster} lists the reduction in overall execution time due to parallelizations introduced. We varied the number of task clusters used for meta training and always created 2 cluster groups to keep degree of cluster training parallelization fixed at 2. This was done due to hardware limitations of our setup. However, the improvement in terms of time saving is encouraging. The experiments were speeded by more than 2x (base time vs optimized time) while the accuracy achieved on test task set remained same.    

\begin{center}
	\begin{table}[ht]
		\small
		\caption{Clustering of tasks for Meta Training.}
		\begin{tabular}{ |p{1.8cm}|p{2.3cm}|p{3cm}|}
			\hline
			Num Clusters & Base Time (in sec)& Optimized Time (in sec)\\ \hline
			2 & 1450.15 & 826.53 \\ \hline
			3 & 6313.8 & 2977.83\\ \hline
			4 & 11361.3 & 4918.31\\ \hline
			5 & 21932.75 & 12473.1 \\ 
			\hline\end{tabular}
		
		\label{table:cluster}
	\end{table}
\end{center}

We also introduced program and code level changes mentioned in Figure \ref{fig:optim_tech}. Next we discuss what changes were made and how they improved the performance.

\subsection{Program and Code Optimizations}
There are several straight forward optimization like upgrading the libraries and platforms used in the available code. For example, upgrading python itself resulted in 15\% speedup. However, to our surprise upgrading Tensorflow Library didn't result in any speedup.

\begin{figure}
	\centering
	\includegraphics[width=.85\linewidth]{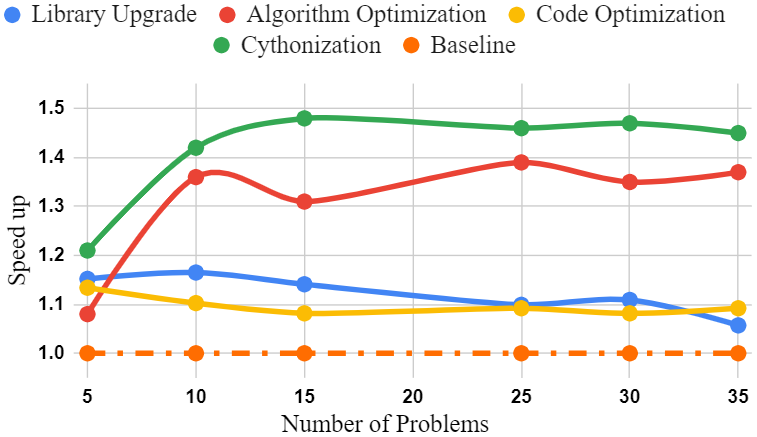}
	\caption{Comparison of trend in speed up for optimization techniques.}
	\label{fig:comparison}
\end{figure}

We also used vector instructions inside of loops using Numpy arrays and removed redundant code by fusing some of the operations performed, put together these Code Optimizations resulted in 1.1X performance improvement. Cythoning the code (converting Python code to C) shows the most promising results with an average speed up of 1.33x. The impact of these individual optimizations is presented as graph in Figure \ref{fig:comparison}. We varied the number of tasks from 5 to 35 as shown in X-axis. This variation resulted in different number of task clusters being formed with different number of tasks in them, thus, covering wider range of scenarios.

\begin{figure}
	\centering
	\includegraphics[width=.85\linewidth]{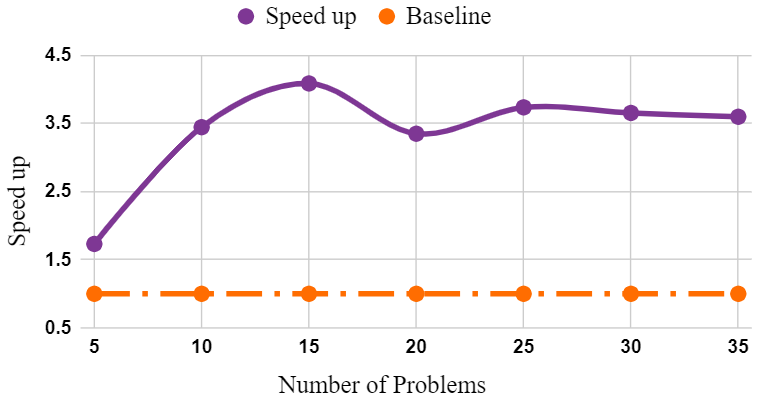}
	\caption{Trend in overall speed up as number of tasks meta trained on are increased.}
	\label{fig:speedup}
\end{figure}

Finally, we combined all the optimizations to get a speed up of 3.73x. Figure \ref{fig:speedup} presents the trend in variation in speed up as the number of tasks being meta trained on are increased. Initially, with lesser number of tasks, the performance curve shows improvement in performance reaching upto 4x and later stabilizing at 3.73x. 

\section{on-going and future work}
\label{sec:ongoing_future}

We are experimenting with different ways and granularities in which the task clusters can be created. We believe grouping similar tasks together in clusters and training meta learner over them will result in faster convergence than without clustering. This is because the the gradient updates for tasks in same cluster will be more similar to each other. 

We are also experimenting whether sequence in which tasks-cluster are used for meta-learning has impact on convergence time of meta learning. A simple analogy from racket sports is that a person good at Racket-ball will find it easier to learn Squash as compared to Tennis \footnote{In Squash, Racket-ball the ball bounces back from the wall and both players are on same court, whereas, in Tennis the courts are different.}.


We also experimented replacing GRU cells used in Hierarchical RNN optitimizer with LSTM in hope of achiving convergence with lesser number of meta iterations. Although, the convergence happened in lesser number of iterations than when GRU cells were used, the time per iteration increased considerably, thus, nullifying the gains. We shelved this idea for future.

\section{conclusions}
\label{sec:conclusions}
We presented our on-going work involving acceleration of a Hierarchical RNN based meta learner proposed in ~\cite{DBLP:conf/icml/WichrowskaMHCDF17}. We changed the sequential meta learning pipeline discussed in paper to a parallel pipeline by introducing parallizations inside training of the tasks as well as across clusters of tasks. We also introduced grouping of tasks clusters for better scheduling.

Apart from introducing wider metalearning pipeline we also introduced certain program and code level optimizations. Put together the proposed techniques resulted in a 3.73X speed-up of the meta-learning algorithm.

\bibliographystyle{ACM-Reference-Format}
\bibliography{biblio}

\end{document}